\documentclass{article} 
\usepackage{collas2022_conference,times}
\usepackage{hyperref}
\usepackage{url}
\usepackage{graphicx}
\usepackage{wrapfig}
\usepackage{float}
\usepackage{booktabs}
\usepackage{subcaption}

\usepackage{amsmath,amsfonts,bm}

\def\eqref#1{equation~\ref{#1}}

\def\1{\bm{1}}

\DeclareMathAlphabet{\mathsfit}{\encodingdefault}{\sfdefault}{m}{sl}
\SetMathAlphabet{\mathsfit}{bold}{\encodingdefault}{\sfdefault}{bx}{n}

\usepackage{hyperref}
\hypersetup{
    colorlinks=true,
    linkcolor=red,
    filecolor=magenta,      
    urlcolor=blue,
    citecolor=purple,
    pdftitle={Reinforcement Learning in a Zipfian World},
    pdfpagemode=FullScreen,
    }

\usepackage{listings}
\title{Zipfian environments for reinforcement learning}

\author{%
Stephanie C. Y. Chan*, Andrew Kyle Lampinen*, Pierre Richemond* \& Felix Hill*\\
*Equal contributions\\
DeepMind\\
London\\
United Kingdom \\
\texttt{\{scychan,lampinen,richemond,felixhill\}@deepmind.com} \\
}

\collasfinalcopy 

\everypar{\looseness=-1}
\begin{document}

\maketitle

\begin{abstract}
As humans and animals learn in the natural world, they encounter distributions of entities, situations and events that are far from uniform. Typically, a relatively small set of experiences are encountered frequently, while many important experiences occur only rarely. The highly-skewed, heavy-tailed nature of reality poses particular learning challenges that humans and animals have met by evolving specialised memory systems. By contrast, most popular RL environments and benchmarks involve approximately uniform variation of properties, objects, situations or tasks. How will RL algorithms perform in worlds (like ours) where the distribution of environment features is far less uniform? To explore this question, we develop three complementary RL environments where the agent's experience varies according to a Zipfian (discrete power law) distribution. These environments are available as an open source library.\footnote{Environments are available at https://github.com/deepmind/zipfian\_environments} On these benchmarks, we find that standard Deep RL architectures and algorithms acquire useful knowledge of common situations and tasks, but fail to adequately learn about rarer ones. To understand this failure better, we explore how different aspects of current approaches may be adjusted to help improve performance on rare events, and show that the RL objective function, the agent's memory system and self-supervised learning objectives can all influence an agent's ability to learn from uncommon experiences. Together, these results show that learning robustly from skewed experience is a critical challenge for applying Deep RL methods beyond simulations or laboratories, and our Zipfian environments provide a basis for measuring future progress towards this goal.
 
\end{abstract}

\section{Introduction}

\begin{wrapfigure}{r}{0.5\textwidth}
  \vspace{-2em}
  \begin{center}
    \includegraphics[width=0.5\textwidth]{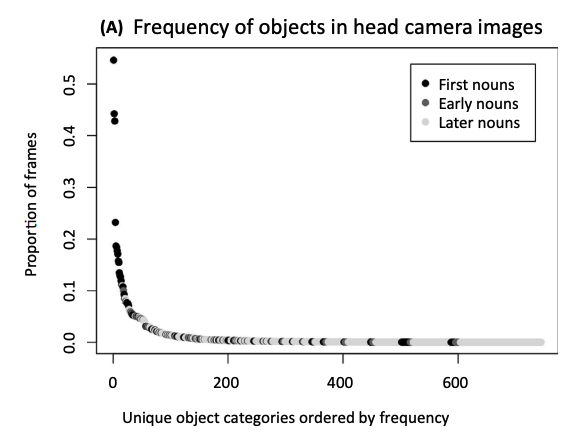}
  \end{center}
\caption{The highly-skewed frequency of common objects (cup, spoon, bowl), as observed in data recorded from 8-10 month old infants wearing head cameras. Reproduced with permission from ~\cite{smith2018developing}.}
\label{fig:smith}
\end{wrapfigure}

Real world distributions are rarely uniform. For example, the words in natural language follow a power-law distribution known as Zipf's law \citep{zipf1936psycho}---frequency is inversely proportional to rank. Thus common words like `the' are extremely frequent, but there is a very long tail of rare words. Similar patterns hold in many other domains of human experience, such as the frequency of objects encountered in early visual learning \citep{clerkin2017real,smith2018developing} --- see Figure~\ref{fig:smith} -- or the relationships we form in social networks \citep{arenas2004community, Albert2001StatisticalMO}. This aspect of natural experience poses a substantial challenge for statistical learning systems, since information about rare entities or events can get lost in a sea of frequent stimuli. Worse, in many domains, rare situations can be critically important. In language, rare words on average have a stronger influence on the meaning of sentences than frequent ones~\citep{bybee1997three}. Similarly, it may be less likely that a self-driving car observes a person rather than a lane marker in the road, but it is more critical to choose appropriate actions around the person.  

Reinforcement Learning (RL) is a powerful paradigm for developing artificial systems that can learn to take optimal decisions from continuous experience of an environment. However, simulated RL environments do not typically expose learners to data that follows a skewed, long-tailed frequency distribution, and as a consequence downplay the value of strong performance in rare situations. Environments that employ random programmatic variation typically vary uniformly \citep{chevalier2018babyai,Cobbe2019QuantifyingGI, Zhang2018ASO, Peng2018SimtoRealTO}. In standard
\begin{wrapfigure}{l}{0.75\textwidth}
  \vspace{-1em}
  \begin{center}
    \includegraphics[width=0.75\textwidth]{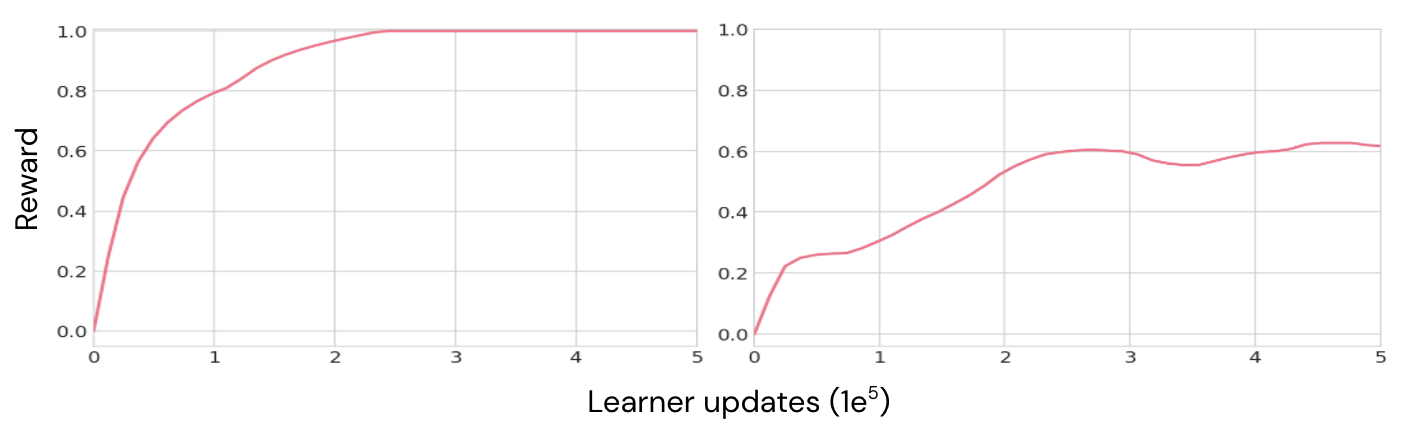}
  \end{center}
\caption{An agent trained with the IMPALA algorithm on in a Zipfian environment (Zipf's Playroom) with 50 objects. \textbf{Left} evaluated on all objects presented according to the training distribution \textbf{Right} on the rarest 20\% of objects in the training distribution.}
\label{fig:misleading}
\end{wrapfigure}

\noindent multi-task settings 
\citep{bellemare2013arcade,brockman2016openai}, 
learners can automatically access equal amounts of experience for each of the multiple tasks. Even in hand-curated game environments, where the distribution of objects and properties is probably not uniform \citep[e.g.][]{vinyals2017starcraft}, it is unlikely that data fully reflects the sharp skew and heavy tails of natural experience. Moreover, irrespective of the distribution of experience from which agents learn, RL agents are typically evaluated under the same frequency distribution as experienced by the learner. Since the agent's score is averaged over the evaluation, performance in high frequency situations is credited most, and failure on rare situations can have little impact on evaluation metrics.

Figure~\ref{fig:misleading} shows how extreme this effect can be: an agent is trained to respond appropriately to one of 50 possible objects presented according to a highly skewed distribution. When evaluated according to the training distribution, its performance looks almost perfect (left), but in fact when it encounters any of the rarest 10 objects, it behaves appropriately less than 60\% of the time (right). Natural intelligence avoids similar failures, thanks to evolved mechanisms like the hippocampus, a rapid learning system that can store and recall individual experiences, complementing a second, slower system of parametric cortical learning \citep{mcclelland1995there}.

Can existing RL algorithms and popular agent architectures allow robust learning about both frequent and rare phenomena? Or will human-like systems for representation and memory be critical, as argued e.g. by \citet{kumaran2016learning}. To assist in resolving these questions, and to highlight the importance of developing systems that can learn from skewed or long-tailed distributions of experience, we develop three new benchmark environments. To parallel natural intelligence, two of these environments --- \emph{Zipf's Playroom} and \emph{Zipf's Labyrinth} --- place agents inside a 3D, first-person simulation. To enable faster experimentation at lower computational cost, the third environment --- \emph{Zipf's Gridworld} --- is a top-down 2D world with a simple discrete action space in which the learner sees (and controls) its own avatar. In Zipf's Playroom, the agent must learn to identify and interact with a global set of 50 different objects which appear according to a heavily-skewed power-law (Zipfian) distribution. In Zipf's Labyrinth, the agent must learn to perform well across a set of 30 tasks when its experience across those tasks is again governed by a Zipfian distribution.  Finally, in Zipf's Gridworld, both the agent's experience of both objects \emph{and} spatial layouts are heavily-skewed.  

Importantly, in all three environments, when evaluating, we measure agent performance under two new distributions. First, to quantify exclusively competence in rare situations, we compute average success when episodes are sampled uniformly across the rarest 20\% of training situations. Second, to verify that performance on common situations is also preserved, we compute average success when episodes are sampled uniformly from \emph{all} possible training situations.  

It is worth noting that RL agents in any setting must learn from experience that is somewhat non-uniform. By their very nature, they must explore, and thus control and change the shape of the data-generating distribution they experience over time - the so-called `generative model' form of exploration \citep{Kearns1999ASS}. Intuitively, reinforcing states with high reward (exploitation) skews the state-visitation probability density, and existing RL algorithms are therefore already designed to handle some agent-induced non-uniformity. Furthermore, many recent RL innovations could potentially adjust for non-uniformity, such as upweighting data that most `surprises' the learner. In particular, Prioritized Experience Replay \citep[PER;][]{Schaul2016PrioritizedER} observes that the empirical distribution of TD-errors in the experience replay buffer of Atari games is heavy-tailed and follows a power law; and that emphasizing surprising transitions promotes better learning. Thus PER could potentially address some of the skew induced by the environment.

However, while these existing methods can learn despite the inevitable \emph{agent-driven} non-uniformity of experience, 
their effectiveness in the face of a \emph{environment-driven} non-uniformity of experience has not been explored. Indeed, it is likely that these two sources of skewed experience interact, and that experience in a highly-skewed environment may actually exacerbate agent-induced non-uniformity of experience. For example, if a particular goal is more frequently rewarding than others, the agent may further bias its experience by consistently pursuing that goal, and thus potentially fail to learn under what conditions the other goals are rewarding.

In an initial exploration of these questions, we evaluate a range of popular RL algorithms and agent architectures on our three Zipfian benchmarks. We focus particularly on three aspects of agent design that may influence performance in rare situations: the \emph{learning objective} (or RL loss), the \emph{memory architecture}, and applications of \emph{self-supervised learning}. We find that, despite apparently `mastering' their training tasks, current popular algorithms in fact perform poorly in rare situations -- in many cases barely above chance. Modified learning objectives, memory architectures and self-supervised auxiliary learning can each lead to moderate improvements in rare situations, but performance is still substantially lower than that achieved when training experience is uniform rather than skewed. Our findings pose an important challenge for future research, which is exemplified in our three new benchmarks -- to create learning systems and RL algorithms that can leverage the plentiful access to frequent experiences provided by natural data and real-world learning contexts to perform successfully, robustly, and safely when faced with rare situations.

\section{Benchmarks for learning from skewed experience}

We develop three benchmark environments in which the training environment layouts, objects, or tasks are experienced according to a Zipfian or power law distribution \citep{Newman2004PowerLP} with exponent $\alpha \in [0, \infty)$. That is, the probability distribution for a random variable $X$ is:
\begin{equation}
    p(X=x) = \frac{1}{Z} \cdot \frac{1}{x^{\alpha}} \label{eq:zipf}
\end{equation}

where \(Z\) is a normalizing constant. $\alpha$ thus determines the degree to which the agent's experience is skewed and long-tailed. A higher value of $\alpha$ increases the difference in frequency between common and rare events, and thus should make learning about rare situations more difficult. Note that a larger support for $X$ will also increase the skew. 

The particular values of $\alpha$ chosen for each environment were intended to create a difficult but not insurmountable learning challenge. However, in both Zipf’s Labyrinth and Gridworld it is possible to change the training distributions quite easily; thus researchers could explore different settings when using the environments. The chosen skew levels are within range of many real-world skew levels \citep[e.g.][]{Newman2004PowerLP,piantadosi_zipfs_2014}.

The three benchmark environments we propose cover a range of settings and paradigms---including 2D and 3D environments, language conditioned or visually-cued tasks, etc. Furthermore, these environments target a diverse range of behaviors, ranging from navigation to instruction-following and 

\subsection{Zipf's Playroom}

\begin{figure*}[t!]
\includegraphics[width=\linewidth]{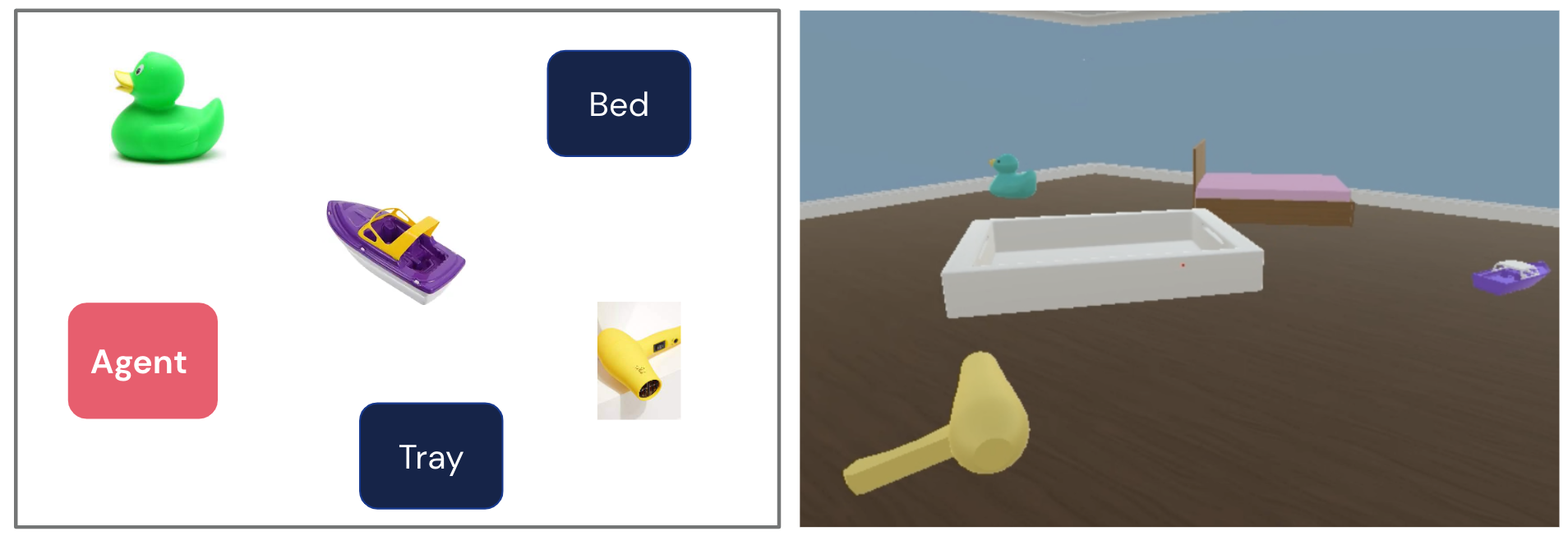}
\caption{Illustration of Zipf's Playroom. \textbf{Left} a schematic of the room from above, showing three randomly chosen toys and the bed and tray furniture. The initial position of each entity and the agent is randomized per episode. The agent would receive an instruction like ``Put a duck on a bed.'' \textbf{Right} A frame of the same environment (with different randomized positions) from the agent's perspective.}
\label{fig:playroom}
\end{figure*}

\emph{Zipf's Playroom} (Figure~\ref{fig:playroom}) is a 3D room built using the Unity game engine. This environment focuses on a highly skewed experience of objects (and their corresponding labels), which parallels the experience of human infants and their highly non-uniform experience of objects and people~\citep{clerkin2017real,smith2018developing}. We consider two tasks: a lifting task and a putting task. In the \textbf{lifting task}, the agent is placed in a room containing three objects. The agent receives a string instruction to lift a specific object. If the agent lifts the correct object, it receives a reward of $1$ and the episode terminates. If the agent lifts an incorrect object, it receives a reward of $0$ and the episode terminates. In the \textbf{putting task}, the agent is placed in a room containing three objects, a red box and a blue box. The agent receives a string instruction to pick up a specific object and to put it into one of the boxes. If the agent puts the correct object into the correct box, it receives a reward of $1$ and the episode terminates. If the agent puts the correct object into the incorrect box or an incorrect object into either box, it receives a reward of $0$ and the episode terminates. 

In both tasks, we sample the three objects that appear in each episode according to a distribution over a total set of 50 objects. During training, these are sampled (independently, without replacement) according to the Zipfian distribution~\ref{eq:zipf} (with exponent $\alpha = 3$ for lifting, and $\alpha = 2$ for putting). However, we evaluate the agent's behaviour when they are sampled from a uniform distribution (\textbf{uniform evaluation)}, or from a uniform distribution truncated to the 10 least frequent items in the Zipfian training distribution (\textbf{rare item evaluation}). Once the three objects are determined, the target object (and hence the instruction) is determined according to a uniform choice between the three sampled objects. In all cases, evaluation consists of running 1000 episodes on the environment tasks, to estimate expected performance under the joint distribution of task-orthogonal and object-sampling randomness. 

In training and testing, we vary several task-orthogonal aspects of the environment randomly according to a uniform (continuous or categorical) distribution: the initial positions and orientations of all objects, boxes and the agent. The variation during training encourages the agent to learn more general policies \citep{Cobbe2019QuantifyingGI}, while the variation at evaluation reduces the possibility that observed performance is due to some idiosyncratic bias.

The putting task additionally allows us to investigate the interaction between long-tailed distributions and systematic or compositional generalization \citep{fodor_connectionism_1988, lake_generalization_2018, hill2019environmental}. Training is performed on a curricula of three tasks that are trained on simultaneously. In order of increasing difficulty: lifting (same as above), putting near (same as the full task except reward is also given if the correct object is placed near the target box), and putting on (the full task). For `putting near' and `putting on', we only trained on the most frequent 20\% of the 50 objects. Thus, to perform well at evaluation on the non-frequent objects, the agent must perform systematic generalization by composing its knowledge of the non-frequent objects with its knowledge of putting (combinations that were never required in training) -- this is an especially difficult setup for systematic generalization, because of diminished experience with the non-frequent objects.

\subsection{Zipf's Labyrinth}

\begin{figure*}[t!]
\includegraphics[width=\linewidth]{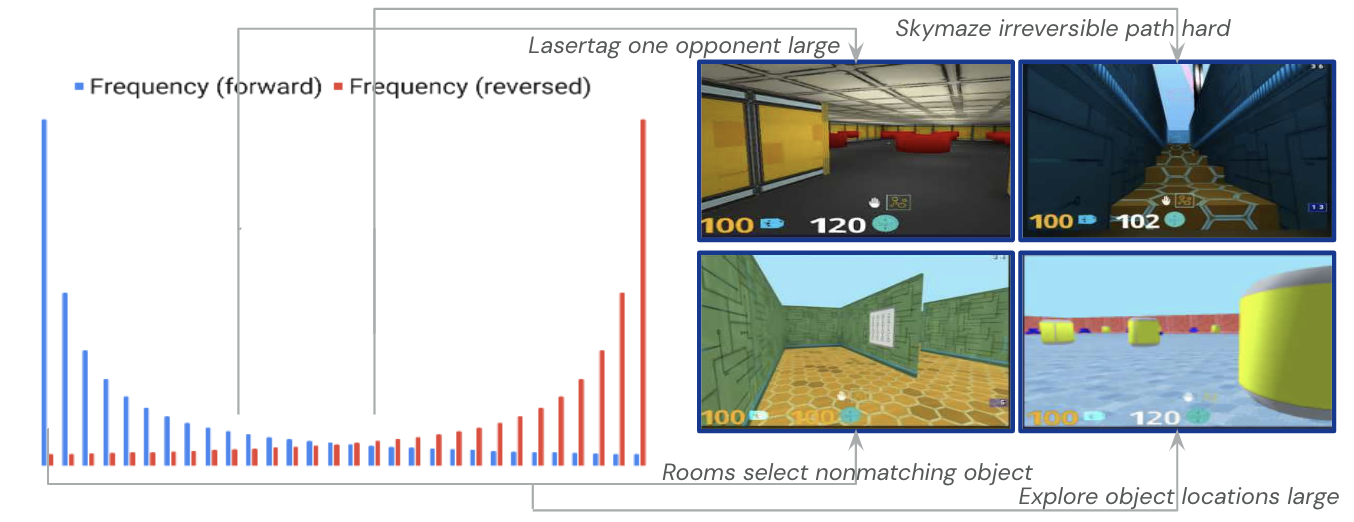}
\caption{\textbf{Left} the frequency with which the learner experiences each of the DM-Lab 30 tasks in Zipf's Labyrinth. \textbf{Right} Frames of several tasks from the agent's perspective.}
\label{fig:labyrinth}
\end{figure*}

While Zipf's Playroom focuses on object identification and manipulation, \emph{Zipf's Labyrinth} (Figure~\ref{fig:labyrinth}) focuses on heavily-skewed experience of tasks and situations that pertain to specific goals. To create it, we simply re-balance the existing \emph{DM-Lab} benchmark \citep{DBLP:journals/corr/BeattieLTWWKLGV16} -- a collection of 30 distinct tasks set in a 3D, first person environment built on Quake 3 Arena.\footnote{\url{https://github.com/id-Software/Quake-III-Arena}} During training, agents experience each task with probability determined by a Zipfian distribution (with exponent 1). We create this Zipfian distribution according to the original ordering of the tasks, which groups the tasks by type. This may amplify the challenge caused by the skew, because entire clusters of similar tasks will be rare or common. To ensure that our results are not solely determined by this particular ordering, we trained both on Zipfian distributions that decreased across the tasks (``forward Zipf'') and also that increased across the tasks (``reversed Zipf''). As above, 
performance is measured across all tasks sampled uniformly, 
and also on the rarest 20\% (i.e. six tasks).

\subsection{Zipf's Gridworld}

\begin{figure}
    \centering
    \includegraphics[width=0.66\textwidth]{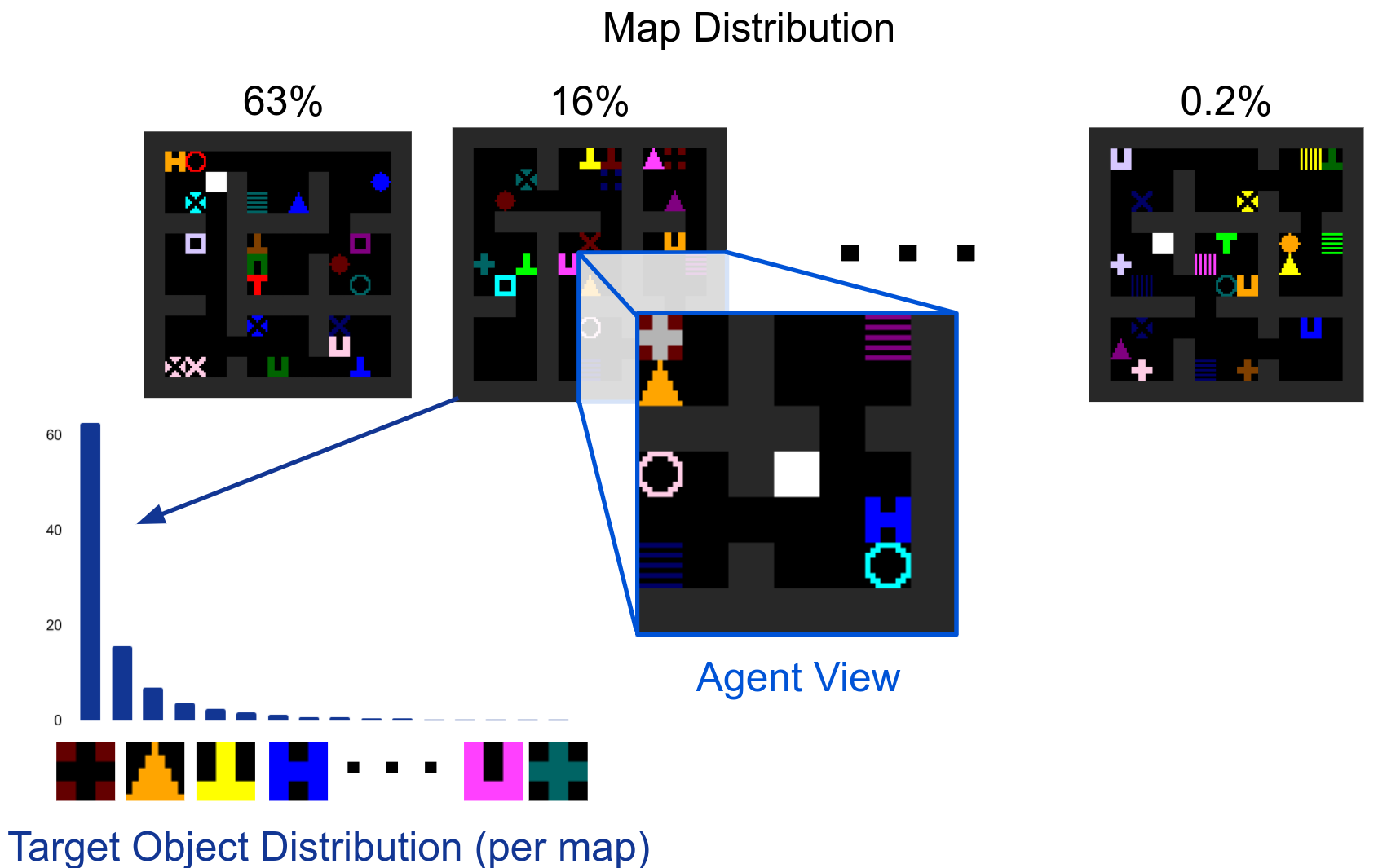}
    \caption{\textbf{Zipf's Gridworld: a 2D world with hierarchical skew.} In each episode, a map is picked from a Zipf distribution over 20 possible maps (top). The agent (white square) and objects always start at the same location within a given map. Each map contains 20 objects, and one of them is chosen to be the target according to a Zipf distribution (bottom left). Thus there is skew across the maps, and skew within each map. The agent sees a local region (bottom right), and the target object is displayed in the top left corner of its view over a light grey highlight background (here, the red collection of squares on the light background is the target object cue).}
    \label{fig:gridworld}
\end{figure}

\emph{Zipf's Gridworld} (Fig. \ref{fig:gridworld}) is a lighter-weight set of tasks for easier experimentation. To make experimenting more accessible, this environment has a simpler action space, shorter episodes, and only a visual input modality. The tasks are visually-cued object finding---the agent has to find the object depicted in the top-left corner of its visual input---in a 2D map containing several rooms and many objects. There are a fixed set of 20 maps, each of which has 9 rooms and 20 objects. The agent start location, as well as the object shapes, colors, and locations are fixed within each map. On each episode, the agent is visually cued to go to a particular target object (see Figure), and rewarded if it moves to that object (the episode ends with no reward if it touches any other object). Because the locations remain consistent within each map, it suffices for the agent to memorize the sequence of actions leading from each map's starting location to each goal object. However, the agent encounters the maps with a Zipfian distribution (with \(\alpha=2\)), so that some are much more common than others, and within each map the goal is also chosen according to a Zipfian distribution. Zipf's Gridworld thus evaluates the agents ability to generalize from hierarchically skewed experience to uniform distributions over maps and goal objects, as well as uniform distributions over the rarest 20\% of each---the 4 rarest objects on the 4 rarest maps.

\section{Reinforcement Learning from rare experiences}

Various aspects of popular (Deep) RL agents could influence their ability to learn effectively from rare experiences. 

\textbf{Prioritized experience replay.} Prioritized experience replay \citep[PER;][]{Schaul2016PrioritizedER} is designed to help replay-buffer-based Deep RL algorithms \citep{Lin1992SelfimprovingRA, mnih2015humanlevelct} by prioritizing `surprising' transitions. Each transition is assigned a priority based on the magnitude of its temporal-difference (TD) error, which is an observable (but imperfect) surrogate for its contribution to learning. Motivated by the empirical finding that the distribution of TD-errors stored in Atari experience buffers approximately follows a power-law distribution, the PER correction samples transitions proportional to a power law over the priority. The ablation study in \citet{Hessel2018RainbowCI} found that PER forms a crucial component of the performance of integrated Deep RL agents. The performance gains provided by similar rebalancing strategies are obvious in supervised learning as well \citep[e.g.][]{Hinton2007ToRS}. Thus, PER's emphasis on surprising experiences, may help compensate for environment-induced skew in experience by increasing the priority of rare events. However, PER 
has not been tested specifically for its effect in skewed environments. This motivates our empirical analysis of the benefit of PER for multi-task settings in Zipfian worlds. 

\textbf{Self-supervised learning.} Self-supervised learning (SSL) \citep{Pathak2016ContextEF, Doersch2017MultitaskSV, Oord2018RepresentationLW, Hnaff2020DataEfficientIR, He2020MomentumCF, Chen2020ASF, Grill2020BootstrapYO, Mitrovic2021RepresentationLV, Caron2021EmergingPI, Radford2021LearningTV} is a family of methods that aim to leverage the inherent structure of the training data in order to learn without supervision. In theory, many SSL methods should lead to grouping similar data points together somewhat irrespectively of the frequency with which those points appear; in fact some self-supervised learning algorithms were derived from (or admit) a \emph{clustering} interpretation \citep{Caron2020UnsupervisedLO, Asano2020SelflabellingVS, HaoChen2021ProvableGF}. Indeed, there is some empirical evidence that SSL can make supervised learners more robust to dataset imbalance \citep{Liu2021SelfsupervisedLI}, and in particular Zipfian shaped or long-tailed  distributions of input data \citep{Zhang2021TestAgnosticLR}. Similarly, difficult and risk-sensitive problems in real world computer vision settings featuring a long tail of open classes such as medical \citep{ ,Ghesu2022SelfsupervisedLF} or autonomous driving data \citep{Mittal2020JustGW} have been addressed using SSL. 

However, the theoretical motivations for self-supervised reconstruction losses in RL are ambiguous, 
due to potential misalignment between the geometry of representations and environment structure \citep{Zhang2021LearningIR}. Tackling this misalignment may require more sophisticated representation-learning approaches (ibid); alternatively, SSL methods may require careful tuning to each environment and task, to accommodate differences in observation-task alignment. Nevertheless, previous work has shown empirically that SSL can improve the robustness and sample-efficiency of RL~\citep{jaderberg2016reinforcement,nair2018visual,Guo2018NeuralPB, Wu2019TheLI, Guo2020BootstrapLR, Schwarzer2021DataEfficientRL, Ye2021MasteringAG}. Thus we evaluate whether SSL can improve robustness of learning in Zipfian environments.

\textbf{Memory systems.} It is also possible that the agent's within-episode memory architecture affects its ability to learn from rare events. In particular, transformers \citep{vaswani2017attentionia} have shown substantial ability to learn about rare events, with the largest language models exhibiting recall of some rare training experiences \citep[e.g.][]{carlini2022quantifying}. While evaluating such a large transformer would be prohibitively difficult, we compare to agents with a smaller transformer-style memory, to evaluate whether the transformer architecture can affect learning from rare experiences. Specifically, we compare to a Gated TransformerXL memory, which has been previously shown more rapid and stable learning in RL than non-gated transformers \citep{Parisotto2020StabilizingTF}. 

\section{Experiments}

We experiment with three popular RL algorithms that exhibit strong performance in 3D first-person environments. Two are policy gradient algorithms --- IMPALA~\citep{Espeholt2018IMPALASD} and V-MPO~\citep{Song2020VMPOOM} --- and one is a recurrent, distributed version of deep Q-learning~\citep{mnih2015humanlevelct} --- R2D2~\citep{kapturowski2018recurrent}. In each case, the agent architecture consists of an \emph{encoder}, for embedding visual (pixel) and language (string) observations from the environment engine, a \emph{memory core} for integrating these observation embeddings over time, and policy, value or Q-value \emph{heads} for mapping the output of the memory core into a representation that can inform behaviour. The encoder consists of a small convolutional network~\citep{Fukushima1980NeocognitronAS} and LSTM networks whose sizes depends on the environment. By default, the memory core is an LSTM~\citep{Hochreiter1997LongSM}, but, we also explore the effect of a Transformer-based memory~\citep{Parisotto2020StabilizingTF}.

To explore the effect of SSL, we employ a generic reconstruction loss \citep[as per][]{hill2020grounded}, in which additional networks are trained, conditioned on the state $h_t$ of the agent's memory, to predict either the current visual ($o_t$ - containing pixel values) or language ($l_t$) observation. This prediction network is either a deconvolutional network or a sequential (LSTM) language decoder, and the prediction loss (a cross-entropy loss, in the case of vision, and a sequence likelihood loss, for language) is backpropagated into the agent's core together with the RL (policy-gradient) loss. We also experiment with a predictive SSL loss that adapts the BYOL algorithm~\citep{Grill2020BootstrapYO} to an RL setting.

Finally, we investigate the benefit of prioritized experience replay in the context of the replay-based R2D2 agent. 

For each condition, we set hyperparameters by manual (informal) search guided only by performance on the skewed training set, not the evaluation environments. We initialized these searches with hyperparameters that have been shown to work well in uniformly-distributed versions of environments similar to ours. When we have found hyperparameters that enable strong performance on the training set, we fix them and train three randomly-initialised agent replicas per condition to convergence. We take the median run across the three runs, and report the median performance in a time window that covers the relatively converged sections of the learning curves. The windows we used were (number of learner updates): [300k, 400k] for Zipf's Playroom, [200k, 380k] for Zipf's Labyrinth, and [2M, 3M] for Zipf's Gridworld. However, evaluating in other similar windows would not have substantially altered the results in most cases, at least in terms of the overall ordering of performance, although performance might continue to slowly improve with longer training. For the Zipf's Labyrinth results, we normalized all results to human-level performance (as reported in \citealt{Espeholt2018IMPALASD}) to better aggregate performance measures across tasks, which can have very different ranges of reward values. All hyperparameters, together with other details of the agent networks are reported in Appendix~\ref{app:hyperparameters}.

\section{Results}
Our results are summarized in Table \ref{tab:playroom_results} for Zipf's Playroom, Table \ref{tab:labyrinth_results} for Zipf's Labyrinth, and Table \ref{tab:gridworld_results} for Zipf's Gridworld. We report training performance for each experiment in Appendix \ref{apps:analyses:train_perf}. Full learning curves are provided in Appendix \ref{apps:learning_curves}.

Across all environments, we see that evaluation performance on rare scenarios is consistently lower than in more frequent situations, sometimes dramatically so. This difference is obscured when we evaluate on the (Zipfian) training distribution, and even (to a lesser extent) when evaluating uniformly across all items, tasks, or maps. Beyond reporting these basic differences in performance in rare situations, our experiments were designed to evaluate whether different architectural and algorithmic components can ameliorate these differences.

\begin{table}[t!]
\begin{tabular}{ll|l|l|l|l|}
                                  &                     & \multicolumn{2}{c|}{Lifting task}                         & \multicolumn{2}{c|}{Putting task}                     \\
                                  &                     & \multicolumn{1}{l|}{All items (uniform)} & Rare items     & \multicolumn{1}{l|}{All items (uniform)} & Rare items \\ \hline
\multicolumn{1}{l|}{RL}           & IMPALA              & 0.65 \tiny{$\pm 0.02$}   & 0.59 \tiny{$\pm 0.02$}   & 0.62 \tiny{$\pm 0.02$}    & 0.49 \tiny{$\pm 0.04$}    \\
\multicolumn{1}{l|}{algorithm}    & V-MPO               & 0.39 \tiny{$\pm 0.01$}   & 0.40 \tiny{$\pm 0.03$}   & 0.00 \tiny{$\pm 0.00$}    & 0.00 \tiny{$\pm 0.00$}    \\
\multicolumn{1}{l|}{}             & R2D2                & 0.57 \tiny{$\pm 0.00$}   & 0.51 \tiny{$\pm 0.00$}   & 0.61 \tiny{$\pm 0.05$}    & 0.44 \tiny{$\pm 0.08$}    \\
\cline{1-2}
\multicolumn{1}{l|}{SSL}          & Vis recon           & 0.66 \tiny{$\pm 0.02$}   & 0.62 \tiny{$\pm 0.06$}   & 0.67 \tiny{$\pm 0.05$}    & 0.55 \tiny{$\pm 0.10$}    \\
\multicolumn{1}{l|}{(IMPALA +..)} & Vis+lang recon      & 0.64 \tiny{$\pm 0.01$}   & 0.66 \tiny{$\pm 0.04$}   & 0.68 \tiny{$\pm 0.01$}    & 0.58 \tiny{$\pm 0.04$}    \\
\multicolumn{1}{l|}{}             & BYOL                & 0.55 \tiny{$\pm 0.01$}   & 0.53 \tiny{$\pm 0.03$}   & 0.62 \tiny{$\pm 0.25$}    & 0.50 \tiny{$\pm 0.18$}    \\ \cline{1-2}
\multicolumn{1}{l|}{Memory}       & MLP                 & 0.47 \tiny{$\pm 0.01$}   & 0.45 \tiny{$\pm 0.02$}   & 0.00 \tiny{$\pm 0.01$}    & 0.00 \tiny{$\pm 0.01$}    \\
\multicolumn{1}{l|}{system}       & LSTM                & 0.65 \tiny{$\pm 0.02$}   & 0.59 \tiny{$\pm 0.02$}   & 0.62 \tiny{$\pm 0.02$}    & 0.49 \tiny{$\pm 0.04$}    \\
\multicolumn{1}{l|}{(IMPALA +..)} & Transformer         & 0.46 \tiny{$\pm 0.05$}   & 0.46 \tiny{$\pm 0.04$}   & 0.59 \tiny{$\pm 0.30$}    & 0.43 \tiny{$\pm 0.24$}    \\
\multicolumn{1}{l|}{}   & Trnsf.+vis+lang recon    & 0.57 \tiny{$\pm 0.00$}   & 0.52 \tiny{$\pm 0.01$}   & 0.50 \tiny{$\pm 0.10$}    & 0.36 \tiny{$\pm 0.08$}    \\
\end{tabular}
\caption{Evaluation performance in \textbf{Zipf's Playroom}, and the effects of different RL algorithms, self-supervised learning objectives and memory systems. We report median  performance across a large evaluation window ([300k, 400k]  updates) for the median of three runs ($\pm$ median absolute deviation across runs). Train performance (assessed on the Zipfian distribution) was generally much higher than evaluation performance, at 0.97-1.00 for lifting and 0.97-0.99 for putting (excluding V-MPO and IMPALA+MLP)---see Appendix \ref{apps:analyses:train_perf} for training results.}
\label{tab:playroom_results}
\end{table}

\begin{table}[t!]
\begin{tabular}{ll|l|l|l|l|l|l|}
                                  &                     & \multicolumn{3}{c|}{Forward Zipf}    & \multicolumn{3}{c|}{Reversed Zipf}     \\  \cline{3-8}
                                  &                     & All tasks     & All tasks     & Rare tasks  & All tasks     & All tasks     & Rare tasks \\
                                  &                     & (Zipfian)     & (uniform)     &             & (Zipfian)     & (uniform)     &            \\  \hline
\multicolumn{1}{l|}{RL}           & IMPALA              & 0.55 \tiny{$\pm 0.26$}  & 0.31 \tiny{$\pm 0.09$}  & 0.32 \tiny{$\pm 0.11$}  & 0.45 \tiny{$\pm 0.10$}  & 0.31 \tiny{$\pm 0.04$}  & 0.35 \tiny{$\pm 0.05$}  \\
\multicolumn{1}{l|}{}             & R2D2                & 0.46 \tiny{$\pm 0.25$}  & 0.26 \tiny{$\pm 0.06$}  & 0.30 \tiny{$\pm 0.11$}  & 0.47 \tiny{$\pm 0.16$}  & 0.31 \tiny{$\pm 0.06$}  & 0.36 \tiny{$\pm 0.14$}  \\
\cline{1-2}
\multicolumn{1}{l|}{SSL}          & Vis recon           & 0.57 \tiny{$\pm 0.22$}  & 0.36 \tiny{$\pm 0.05$}  & 0.35 \tiny{$\pm 0.07$}  & 0.46 \tiny{$\pm 0.09$}  & 0.31 \tiny{$\pm 0.04$}  & 0.34 \tiny{$\pm 0.07$}  \\
\multicolumn{1}{l|}{(IMPALA}      & Vis+lang recon      & 0.58 \tiny{$\pm 0.25$}  & 0.37 \tiny{$\pm 0.07$}  & 0.34 \tiny{$\pm 0.09$}  & 0.45 \tiny{$\pm 0.13$}  & 0.30 \tiny{$\pm 0.06$}  & 0.36 \tiny{$\pm 0.13$}  \\
\multicolumn{1}{l|}{ +..)}        & BYOL                & 0.53 \tiny{$\pm 0.26$}  & 0.29 \tiny{$\pm 0.08$}  & 0.27 \tiny{$\pm 0.10$}  & 0.05 \tiny{$\pm 0.01$}  & 0.05 \tiny{$\pm 0.01$}  & 0.07 \tiny{$\pm 0.03$}  \\
\cline{1-2}
\multicolumn{1}{l|}{Memory}       & MLP                 & 0.40 \tiny{$\pm 0.24$}  & 0.17 \tiny{$\pm 0.05$}  & 0.14 \tiny{$\pm 0.05$}  & 0.04 \tiny{$\pm 0.01$}  & 0.05 \tiny{$\pm 0.01$}  & 0.07 \tiny{$\pm 0.03$}  \\
\multicolumn{1}{l|}{system}           & LSTM              & 0.55 \tiny{$\pm 0.26$}  & 0.31 \tiny{$\pm 0.09$}  & 0.32 \tiny{$\pm 0.11$}  & 0.45 \tiny{$\pm 0.10$}  & 0.31 \tiny{$\pm 0.04$}  & 0.35 \tiny{$\pm 0.05$}  \\
\multicolumn{1}{l|}{(IMPALA} & Transformer              & 0.46 \tiny{$\pm 0.25$}  & 0.19 \tiny{$\pm 0.06$}  & 0.05 \tiny{$\pm 0.01$}  & 0.35 \tiny{$\pm 0.11$}  & 0.22 \tiny{$\pm 0.05$}  & 0.22 \tiny{$\pm 0.06$}  \\
\multicolumn{1}{l|}{ +..)}   & Trnsf.+vis+lang recon    & 0.55 \tiny{$\pm 0.26$}  & 0.31 \tiny{$\pm 0.09$}  & 0.29 \tiny{$\pm 0.10$}  & 0.37 \tiny{$\pm 0.11$}  & 0.28 \tiny{$\pm 0.06$}  & 0.36 \tiny{$\pm 0.16$}  \\
\end{tabular}
\caption{Evaluation performance in \textbf{Zipf's Labyrinth}, and the effects of different RL algorithms, self-supervised learning objectives and memory systems. We report median performance across a large evaluation window ([200k, 380k] updates) for the median of three runs, and an average or weighted average across tasks ($\pm$ standard error across tasks). Training performance was consistently higher, as shown in the column ``All tasks (Zipfian)'', compared to evaluation uniformly across all tasks, or on rare tasks only.}
\label{tab:labyrinth_results}
\end{table}

\begin{table}[t!]
\begin{tabular}{ll|l|l|}
                                  &                       & All maps and objects (uniform) & Rare maps and objects        \\ \hline
\multicolumn{1}{l|}{RL}           & IMPALA                & 0.69 \tiny{$\pm 0.05$}        & 0.24 \tiny{$\pm 0.04$}      \\
\multicolumn{1}{l|}{}             & R2D2                  & 0.35 \tiny{$\pm 0.06$}        & 0.18 \tiny{$\pm 0.01$}      \\
\cline{1-2}
\multicolumn{1}{l|}{SSL}          & Vis recon              & 0.80 \tiny{$\pm 0.03$}        & 0.29 \tiny{$\pm 0.07$}      \\
\multicolumn{1}{l|}{}             & BYOL                   & 0.31 \tiny{$\pm 0.02$}        & 0.03 \tiny{$\pm 0.01$}      \\ \cline{1-2}
\multicolumn{1}{l|}{Memory}       & MLP                    & 0.77 \tiny{$\pm 0.01$}        & 0.13 \tiny{$\pm 0.06$}       \\
\multicolumn{1}{l|}{system}       & LSTM                   & 0.69 \tiny{$\pm 0.05$}        & 0.24 \tiny{$\pm 0.04$}      \\
\multicolumn{1}{l|}{(IMPALA +..)} & Transformer            & 0.82 \tiny{$\pm 0.01$}        & 0.22 \tiny{$\pm 0.02$}      \\
\multicolumn{1}{l|}{} & Transformer+vis+lang recon         & 0.70 \tiny{$\pm 0.01$}        & 0.23 \tiny{$\pm 0.04$}      \\
\end{tabular}
\caption{Evaluation performance in \textbf{Zipf's Gridworld}, and the effects of different RL algorithms, self-supervised learning objectives and memory systems. We report median performance across a large evaluation window ([2M, 3M] updates) for the median of three runs ($\pm$ median absolute deviation across runs). Train performance (assessed on a Zipfian distribution) was generally much higher than evaluation performance, at 0.85 for IMPALA and 0.99 for IMPALA + Vis recon---see Appendix \ref{apps:analyses:train_perf} for detailed training results.}
\label{tab:gridworld_results}
\end{table}

\textbf{Does prioritized experience replay improve performance in rare situations?} While on some tasks, training performance of the R2D2 algorithm was lower than our baseline IMPALA agent, we did find that PER made a difference to the underlying Q-learning approach on rare items. R2D2 without prioritized experience replay performed $17\%$ lower on Zipf's Playroom putting tasks ($0.51$ average on all items, $0.36$ on rare items) than the same algorithm using PER. We tried increasing the prioritization exponent in an effort to handle the more extreme skew in our environments, but we found it made little difference.

\textbf{Does self-supervised learning improve performance in rare situations?} 
Comparing the baseline IMPALA agent against IMPALA combined with various SSL algorithms, we see some moderate signs that self-supervised learning could improve agent performance in rare situations. In Zipf's Playroom, adding a vision+language based reconstruction loss to the IMPALA policy gradient loss improves performance by $+12\%$ on rare items in the lifting task and $+18\%$ on the putting task, with vision and language contributing approximately equal amounts to this uplift. These improvements are beyond the benefits of SSL more generally, vs. for the rare items specifically, e.g. as measured by the change in uniform-all performance ($-1\%$ for lifting and  $+10\%$ for putting). In Zipf's Gridworld, where language is not part of the environment, the equivalent visual SSL yields a $+20\%$ improvement on the rarest situations (and $+16\%$ when evaluated uniformly across all maps and objects. It is notable that both Zipf's Playroom and Zipf's Gridworld emphasise the agent's ability to distinguish objects of different categories, which is comparatively less important in Zipf's Labyrinth. This may explain why the benefit of SSL was lower there ($+3-6\%$). Note also that our approach based on BYOL failed, possibly because for simplicity, we did not enforce temporal consistency of our representations, or specifically tune an augmentations set to the problem.

\textbf{Do memory architectures improve performance in rare situations?}
While transformer architectures have enabled large language models to improve performance on rare experiences \citep{carlini2022quantifying}, we saw little evidence that replacing the agent's core memory with a transformer (instead of an LSTM or MLP) led to better performance on rare items. There are many differences in the experience, objective, and scale that could potentially explain this difference---for example modern language models can condition on words across many consecutive sentences, while the IMPALA and V-MPO algorithms do not enable an agent to condition on stimuli outside of the current episode. This motivates research into systems that allow agents to take decisions based on (memories of) extra-episodic as well as episodic context. We also did not find

\textbf{Do RL agents or supervised learners perform better on rare stimuli?} As noted previously, RL agents must contend with both agent-induced and environment-induced non-uniformity when learning from their experience. In contrast, supervised learners only contend with environment- (or data-) induced non-uniformity. We might then expect that --- everything else being equal --- a supervised learning system would perform better in rare situations than an RL agent. To test this hypothesis, we created a bespoke `finding' task in Zipf's Playroom. In this task, the agent spawns facing two objects, one on the left and one on the right, randomly chosen according to a Zipfian distribution as before. We then trained an RL agent to approach a specific object given its name. At the same time, we trained a supervised classifier with the same encoders as the IMPALA agent to predict either \emph{left} or \emph{right} given the first-frame of an episode, via a cross-entropy loss. At test time, we evaluated the RL agent and the supervised learner, as before, according to uniform distributions on all items and on rare items. If, by chance, both objects in the room were the same, either outputs \emph{left} or \emph{right} were credited it as a correct prediction (as would be the case for the RL agent). 

Contrary to our expectations, we found that the RL agent and supervised classifier performed almost identically (Table~\ref{tab:playroom_results:supervised}). In fact, if the learning curves are plotted as a function of the number of learner updates, the median learning curves are very similar. One possible explanation for this is that, while the reinforcement learner has to contend with agent-induced non-uniformity of experience, its embodiment also affords it access to much more data about rare items (in the form of views from different distances and angles). As has been discussed in prior work~\citep{smith2017developmental,hill2019environmental}, this data might afford RL agents with uniquely rich representations which, in this case, may make up for the greater non-uniformity faced by the RL agent.    

We note that our results only show that existing RL methods cannot \emph{consistently} handle long-tailed distributions; they do not show that these methods can never handle long-tailed distributions at all. We also note that, given the statistical power of the experiments, we cannot over-interpret the differences between RL baselines; the goal here was not to give precise measures for each of these baselines.

\begin{table}[]
\begin{tabular}{ll|l|l|l|}
                                            & All items (Zipfian) & All items (uniform) & Rare items \\ \hline
\multicolumn{1}{l|}{RL}                     & 1.00 \tiny{$\pm 0.00$}  & 0.78 \tiny{$\pm 0.01$}  & 0.66 \tiny{$\pm 0.02$}  \\
\multicolumn{1}{l|}{Supervised classifier}  & 1.00 \tiny{$\pm 0.00$}  & 0.79 \tiny{$\pm 0.00$}  & 0.64 \tiny{$\pm 0.02$}  \\
\end{tabular}
\caption{Comparison between an RL agent (IMPALA + LSTM) and a maximum-likelihood supervised classifier, on an object identification `finding' task in Zipf's Playroom. We report median performance across a large evaluation window ([2M, 3M] updates) for the median of three runs  ($\pm$ median absolute deviation across runs).}
\label{tab:playroom_results:supervised}
\end{table}

\subsection{Future directions}

These empirical observations motivate several possible directions for future research. 

\textbf{Reweighting:} While reweighting may be part of a general solution to the problem of learning from heavily-skewed experience, such a solution should not rely on knowing in what ways the distribution is skewed---in real-world applications, it might be difficult to infer. For example, manually reweighting the tasks from Zipf's Labyrinth would not be a satisfying solution to our benchmark. Instead, reweighting schemes should rely on statistics that the agent can observe---such as TD-error---or on clustering based on the agent's representations.

\textbf{Distributional RL:} Another avenue worthy of exploration is \emph{distributional} reinforcement learning \citep{Bellemare2017ADP, Dabney2018ImplicitQN, Dabney2018DistributionalRL}. Because such algorithms learn a full distribution of state-value functions rather than a single scalar, they may be more suited to handling extremely rare events and rewards that would otherwise be subsumed by the computation of a single expectation.

\textbf{Exploration strategies:} One role of explicit exploration strategies---and notions such as \emph{intrinsic motivation}---is to counterbalance non-uniformity of agent experience by visiting states as uniformly as possible. In a bandit or tabular setting, provably efficient exploration strategies exist \citep{Brafman2002RMAXA, Lattimore2014NearoptimalPB, Azar2017MinimaxRB}, usually relying on the principle of optimism in the face of uncertainty. In the neural network approximation case, explicit state visitation counts are unavailable, and proxies or bonuses \citep{Bellemare2016UnifyingCE, Ostrovski2017CountBasedEW, Burda2019ExplorationBR} must be employed. In an environment in which some objects or situations are much more frequent than others, exploration strategies might encourage agents to prefer those that are less frequent, and could thus play a role in improved performance on our Zipfian benchmarks. At the opposite end of the exploration spectrum (when it is not possible), distributional shift and generalization to unfamiliar states are also particularly relevant issues in \emph{offline RL} \citep{Levine2020OfflineRL}.



\textbf{Explicit recall of episodic memories:}
\citet{kumaran2016learning} argue that natural intelligence relies on complementary learning systems, combining slower weight-based learning with episodic memory of experiences in the hippocampus; Kumaran et al. emphasize that these complementary systems ``may allow the general statistics of the environment to be circumvented by reweighting experiences,'' thus allowing intelligent agents to learn rare-but-important things. Specifically, episodic memories can either allow preferential replay of important memories (as we explored with PER), or allow explicit recall of the memory when it is relevant. While a variety of memories have been proposed that allow RL agents to explicitly recall past experiences, these mostly are applied within a single episode \citep[e.g.][]{wayne2018unsupervised,Parisotto2020StabilizingTF}, or across at most a small number of tasks or episodes \citep{ritter2020rapid,lampinen2021towards}. It is unclear whether these methods could be scaled to an entire lifetime of memory---as would be needed to handle Zipfian skew or other similar data distributions. Thus biologically motivated alternatives capable of explicitly recalling specific events at evaluation time, such as episodic control \citep{Lengyel2007HippocampalCT,Blundell2016ModelFreeEC, pritzel2017neural}, may provide an exciting direction for future investigation.



\section{Related Work}

The interplay of rare events and RL, the explicit focus of the present work, was previously studied in the tabular setting with linear evaluation in  \citet{Frank2008ReinforcementLI}, where importance sampling methods were devised to reduce variance. To our knowledge, however, no works directly investigate performance on rare states for Deep RL algorithms trained on an explicitly Zipfian distribution of states or environments.

A number of studies in computer vision have investigated the challenge of learning from skewed training data in supervised settings, e.g. on objects~\citep{zhu2014capturing,clerkin2017real} or faces~\citep{liu2015deep}. This challenge has been formalised in \emph{Long-tailed ImageNet}, a computer vision benchmark in which training data is heavily-skewed \citep{Liu2019LargeScaleLR}. Our Zipfian environments provide an analogous test for embodied reinforcement learning agents. 

In such supervised settings, it is common to simply re-sample or re-weight classes so that the model is exposed equally to all classes \citet{van_hulse_experimental_2007}; some more sophisticated methods e.g. produce artificial minority samples \citep{chawla_smote_2002}. For evaluation, metrics such as balanced accuracy or the F-measure are commonly used to account for imbalanced classes \citep{johnson_survey_2019}.  However, these metrics and reweighting techniques are not applicable when the non-uniformity is hidden within classes---a common phenomenon known as `hidden stratification'---which can have serious impacts on real-world applications such as medical imaging \citep{oakden-rayner_hidden_2019}, and which some works have attempted to address \citep{sohoni_no_2020, sagawa_distributionally_2020}.

In RL, by contrast, agents are typically evaluated on the same problem they are trained on. Therefore achieving reasonable performance on evaluations that differ from the training distribution (as in our proposed benchmarks) remains an open and challenging problem \citep{Zhang2018ASO, Hessel2019MultitaskDR, Cobbe2019QuantifyingGI, Kirk2021ASO}. For instance, \citep{Cobbe2019QuantifyingGI} find that training on dozens of thousands of levels is required to bridge the generalization gap in their multi-task setting, even when using just a uniform distribution of levels; our Zipfian case is likely to exacerbate the issue. The matter is also complicated by concerns around statistical significance of evaluation \citep{chan2020measuring}.






\section{Conclusions}

The structure of the real world is not uniform. Instead, the world is full of Zipfian distributions in which some entities, objects, or tasks are very common, but most are rare. We have highlighted the challenge this poses for reinforcement learning, and have proposed three diverse benchmark environments in which to investigate various aspects of this challenge---Zipf's Playroom, Zipf's Labyrinth, and Zipf's Gridworld. Across all three settings, we find that achieving high performance on the long tail is difficult. We investigate a variety of solutions, but find only modest, inconsistent benefits to performance on the rarest items. We therefore propose these benchmarks to the community as a challenging evaluation of learning from rare experiences. We hope that our benchmark environments will encourage the development of new RL methods and memory systems, and ultimately the development of agents that can learn from a lifetime of non-uniform experience. 
\bibliography{collas2022_conference}
\bibliographystyle{collas2022_conference}

\appendix
\section{Appendix}

\subsection{Zipfian distributions}

The following python code snippet generates Zipfian distributions for a specified exponent:

\begin{lstlisting}
import numpy

def zipfian_dist(num_entities, exponent=1.):
  vals = 1./(numpy.arange(1, num_entities + 1))**exponent
  return vals / numpy.sum(vals)
\end{lstlisting}

\subsection{Using the benchmarks} \label{app:using_the_benchmark}

\subsubsection{General principles for training and evaluation}
In each of our environments, it is necessary to create separate copies of the environment for each training level and testing level in a given split (see below). 

To evaluate, you should run at least three seeds (independent runs) per condition and evaluate after training has converged to relatively stable performance. You should report average performance of the median seed(s) over a relatively large set of episodes (as a rule of thumb, at least 1000 episodes per evaluation level).

While we encourage exploring variations on our benchmarks (for example increasing or decreasing the Zipf exponent), to avoid confusion we ask that you state any modifications very clearly in your publications or reports, and do not use unqualified versions of our split names when describing your results.

\subsection{Installing and running the benchmarks}
Our benchmarks are released online at \url{redacted}

Once you have cloned the repository, you can install most the needed dependencies for all three benchmarks by running the following commands:
\begin{lstlisting}[language=bash]
python3 -m venv zipf
source zipf/bin/activate
pip install --upgrade pip
pip install -r requirements.txt
\end{lstlisting}
However, note that this will install dependencies for all three benchmarks, and will not install some dependencies which are not listed in pip, see below.

\subsubsection{Zipf's Playroom}
To use Zipf's playroom, you will also need to download and install Docker. Once you have done so, you can create an environment as follows:
\begin{lstlisting}
import zipfs_playroom
env_settings = zipfs_playroom.EnvironmentSettings(
  seed=1, level_name='lift/lift_shape_zipf3')
env = zipfs_playroom.load_from_docker(
  name=FLAGS.docker_image_name, settings=env_settings)
obs = env.reset()
\end{lstlisting}
Note that you will have to instantiate different environments for training and test level (and train on several distinct levels for the `put' tasks), see Table \ref{tab:using_environments:playroom_levels}.

\begin{table}[H]
\centering
\begin{tabular}{l|ll}
 \textbf{Task split} & \textbf{Train levels} & \textbf{Test levels} \\ \hline
 Lifting task & \verb|'lift/lift_shape_zipf3'| & \verb|'lift/lift_shape_uniform'|\\
  &  & \verb|'lift/lift_shape_uniform_rare'|\\ \hline
 Putting task & \verb|'put/lift_shape_zipf2'| & \verb|'put/put_on_bed_tray_all'|\\
  & \verb|'put/put_near_bed_tray_frequent'| & \verb|'put/put_on_bed_tray_rare'|\\ 
  & \verb|'put/put_on_bed_tray_frequent'| & \\ \hline
\end{tabular}
\caption{Train and test level splits in Zipf's Playroom.}
\label{tab:using_environments:playroom_levels}
\end{table}

\subsubsection{Zipf's Labyrinth}
To use Zipf's Labyrinth, you will also need to install DM-Lab. Once you have done so, you can instantiate an environment as follows:
\begin{lstlisting}
import zipfs_labyrinth
env = zipfs_labyrinth(level_name='forward_zipf')
obs = env.reset()
\end{lstlisting}
Note that you will have to instantiate different environments for each training and testing level, see Table \ref{tab:using_environments:labyrinth_levels}.

\begin{table}[h!]
\centering
\begin{tabular}{l|ll}
 \textbf{Task split} & \textbf{Train levels} & \textbf{Test levels} \\ \hline
 Forward Zipf & \verb|'forward_zipf'| & \verb|'uniform'|\\
  &  & \verb|'forward_rare'|\\ \hline
 Reversed Zipf & \verb|'reversed_zipf'| & \verb|'uniform'|\\
  &  & \verb|'reversed_rare'|\\ \hline
\end{tabular}
\caption{Train and test level splits in Zipf's Labyrinth.}
\label{tab:using_environments:labyrinth_levels}
\end{table}

\subsubsection{Zipf's Gridworld}
To use Zipf's Gridworld, you can create an 

\begin{lstlisting}
import zipfs_gridworld 
env = zipfs_gridworld.simple_builder(level_name='zipf_2')
obs = env.reset()
obs2 = env.step(6)
\end{lstlisting}
Note that you will have to instantiate different environments for each training and testing level, see Table \ref{tab:using_environments:gridworld_levels}.
\begin{table}[h!]
\centering
\begin{tabular}{l|ll}
 \textbf{Task split} & \textbf{Train levels} & \textbf{Test levels} \\ \hline
 Zipf 2& \verb|'zipf_2'| & \verb|'uniform'|\\
  &  & \verb|'rare'|\\ \hline
\end{tabular}
\caption{Train and test level splits in Zipf's Gridworld.}
\label{tab:using_environments:gridworld_levels}
\end{table}

\subsection{Environment details}

\subsubsection{Zipf's Playroom}

Zipf's Playroom is implemented using Unity. The agent receives a visual observation of \(96 \times 72 \times 3\) pixel RGB images, and a language observation that is tokenized at the word-level. The agent exerts relatively low-level control, using a discrete action space of 46 actions that allow several different magnitudes of movement forward and back, strafing side-to-side, rotating left and right and looking up and down, and grabbing and manipulating held objects. The physics is relatively realistic, for example the agent has momentum in its movements. The agent acts at 7.5 actions per environment second (30 environment FPS, with 4 action repeats between agent steps). Zipf's playroom runs at around 900 FPS (clock time), comparable to DMLab when running on comparable hardware (an NVIDIA Quadro K600 GPU).

\subsubsection{Zipf's Labyrinth}

This environment was used as in \citet{DBLP:journals/corr/BeattieLTWWKLGV16}, except that the distribution of levels was Zipfian. For further details, see the original publication. The environments run at about 700-1000 FPS (depending on the level) on a  Linux desktop with a 6-core Intel Xeon 3.50GHz
CPU and an NVIDIA Quadro K600 GPU (per the original publication). The Zipf distributions were applied to the levels in the original order:
\begin{lstlisting}
LEVELS_DMLAB30_FORWARD = [
    'rooms_collect_good_objects_train', 'rooms_exploit_deferred_effects_train',
    'rooms_select_nonmatching_object', 'rooms_watermaze',
    'rooms_keys_doors_puzzle', 'language_select_described_object',
    'language_select_located_object', 'language_execute_random_task',
    'language_answer_quantitative_question', 'lasertag_one_opponent_small',
    'lasertag_three_opponents_small', 'lasertag_one_opponent_large',
    'lasertag_three_opponents_large', 'natlab_fixed_large_map',
    'natlab_varying_map_regrowth', 'natlab_varying_map_randomized',
    'skymaze_irreversible_path_hard_v2', 'skymaze_irreversible_path_varied_v2',
    'psychlab_arbitrary_visuomotor_mapping', 'psychlab_continuous_recognition',
    'psychlab_sequential_comparison', 'psychlab_visual_search',
    'explore_object_locations_small', 'explore_object_locations_large',
    'explore_obstructed_goals_small', 'explore_obstructed_goals_large',
    'explore_goal_locations_small', 'explore_goal_locations_large',
    'explore_object_rewards_few', 'explore_object_rewards_many'
]
# Reverse distribution
LEVELS_REVERSED = list(reversed(LEVELS_DMLAB30_FORWARD))
\end{lstlisting}
Note that this ordering clusters the tasks by type, which adds a degree of hierarchy to the skewed experience.

\subsubsection{Zipf's Gridworld}
The agent observes \(7 \times 7\) grid squares around it, rendered at a \(9 \times 9\) pixel resolution pre square, for a total visual observation size of \(63 \times 63\) pixels. The top left square of this observation is replaced with a heads-up display showing the target object, displayed on a distinctive light-gray background. The agent can move to any of the 8 adjacent squares, including diagonals. If it move onto an object, the episode immediately ends, and the agent is rewarded if the object it moved to is the target object. Otherwise, it is not rewarded. The episode also terminates with 0 reward if the agent has not touched an object within 100 steps (although this is quite unlikely even under a random policy, as the object density is high).

The objects for each map are created by combining the following 15 colors and 15 shapes, subject to the constraint that all objects are distinct along at least one dimension within each map (but no constraints across maps):
\begin{lstlisting}
COLORS = [
    "red", "green", "blue", "purple", "orange",
    "yellow", "brown", "pink", "cyan", "dark_green",
    "dark_red", "dark_blue", "teal", "lavender", "rose"
]
SHAPES = [
    "triangle", "empty_square", "plus", "inverse_plus", "ex", 
    "inverse_ex", "circle", "empty_circle", "tee", "upside_down_tee",
    "h", "u", "upside_down_u", "vertical_stripes", "horizontal_stripes"
]
\end{lstlisting}
Objects are placed such that the agent can navigate between any two points in the map without touching an object, to ensure that all rooms and objects are reachable. 

Zipf's Gridworld is implemented using the pycolab gridworld engine \url{https://github.com/deepmind/pycolab}. In a simple benchmark (stepping the environment with a fixed action), we measured that Zipf's Gridworld runs at around 4,700 FPS in a single thread on an Intel Xeon W-2135 CPU.

\subsection{Experiment hyperparameters} \label{app:hyperparameters}

\begin{table}[h!]
\begin{tabular}{@{}ll@{}}
\toprule
All agents                      &                       \\ \midrule
language encoder type           & LSTM                  \\
language encoder embedding size & 32                    \\
language encoder hidden size    & 32                    \\
vision encoder type             & ResNet                \\
vision encoder output channels  & (16, 32, 32)          \\
vision encoder resnet blocks    & (2, 2, 2)             \\
encoder mixing operation        & flatten+concat        \\
                                &                       \\ \midrule
LSTM memory core                &                       \\ \midrule
hidden size                     & 512                   \\
                                &                       \\ \midrule
Transformer memory core         &                       \\ \midrule
d\_model                        & 512                   \\
num layers                      & 4                     \\
num attention heads             & 8                     \\
                                &                       \\ \midrule
Self-supervised learning        &                       \\ \midrule
language reconstruction network & LSTM                  \\
visual reconstruction network   & deconv network        \\
lang recon hyper params         & as encoder            \\
visual recon hyper params       & as encoder            \\
lang recon loss                 & cross-entropy         \\
visual recon loss               & sigmoid cross-entropy \\
BYOL forward hidden             & 100                   \\
BYOL projection hidden          & 200                   \\
BYOL prediction hidden          & 100                   \\
BYOL projection size            & 50                    \\ \bottomrule
\end{tabular}
\caption{Architecture hyperparameters for all agents.}
\label{tab:my-table}
\end{table}

\subsubsection{IMPALA agent}

$
\begin{array}{lrrr}
\hline & \text { Zipf's Labyrinth } & \text { Zipf's Labyrinth } & \text { Zipf's Gridworld } \\
\hline \text { Image Width } & 96 & 96 & 96 \\
\text { Image Height } & 72 & 72 & 72 \\
\text { Action Repeats } & 4 & 4 & 4 \\
\text { Unroll Length } & 128 & 128 & 128 \\
\text { Discount }(\gamma) & 0.99 & 0.99 & 0.99 \\
\text { Baseline loss scaling } & 0.6 & 0.5 & 0.59 \\
\text { Entropy cost } & 1e-4 & 0.01 & 9.4e-5 \\
\text { Action Repeats } & 4 & 4 & 4 \\
\text { Optimizer } & \text { Adam } & \text { Adam } & \text { Adam } \\
\text { Learning rate} & 3e-4 & 1e-4 & 3e-4 \\
\hline
\end{array}
$

\subsubsection{V-MPO agent}

We note that V-MPO hyperparameters may not be optimally tuned for these settings, as it was unable to achieve high performance on the putting tasks, even in training (thus we have omitted those results). While V-MPO achieved relatively high training performance on the lifting tasks, it was still noticeably worse than the other algorithms. However, we were unable to find better hyperparameters for it within the sweeps we considered. Nevertheless, the comparison between V-MPO and the other algorithms should be interpreted cautiously.

$
\begin{aligned}
&\begin{array}{lr}
\hline \text { Parameter} & \text { Value} \\
\hline \text { Agent discount } & 0.995 \\
\text { Image width } & 96 \\
\text { Image height } & 72 \\
\text { Number of action repeats } & 4 \\
\text { Number of LSTM layers } & 2 \\
\text { MPO learning rate} & 3 * 10^{-3} \\
\text { Net learning rate } & 3 * 10^{-5} \\
T_{\text {target }} & 0.5 \\
\epsilon_{\eta} & 0.1 \\
\epsilon_{\alpha} & 0.1 \\
\hline
\end{array}\\
&\text { Table 5: Settings for V-MPO }
\end{aligned}
$

\subsubsection{R2D2 agent}

\subsubsection{Zipf's Playroom}

In \emph{Zipf's Playroom}, there is a distinction between hyperparameters on the \emph{lifting} tasks and the \emph{putting} tasks. \\

$
\begin{array}{l|cc}
\hline \text { Tasks type } & \text { Lifting tasks } & \text {Putting tasks } \\
\hline \text { Number of actors } & \multicolumn{2}{c}{256} \\
\text { Sequence length } &  \multicolumn{2}{c}{128 ( \text {prefix of } l=20 \text { burn-in})} \\
\text { Replay buffer size } & \multicolumn{2}{c}{10^{5}} \\
\text { Minibatch size } & \multicolumn{2}{c}{32} \\
\text { Importance sampling exponent } & \multicolumn{2}{c}{0.6} \\
\text { Priority weight (max) } & 0.5 & 0.9\\
\text { Exploration } \varepsilon & 0.125 & 0.15\\
\text { Discount } \gamma & \multicolumn{2}{c}{0.99} \\
\text { Lambda-returns } \lambda & \multicolumn{2}{c}{0.8} \\
\text { Optimizer } & \multicolumn{2}{c}{\text { AdamW }(\text {Loschilov \& Hutter, 2017) }} \\
\text { Optimizer learning rate } & 5*10^{-4} &5*10^{-5} \\
\text { Optimizer } \varepsilon & \multicolumn{2}{c}{1.25*10^{-6}} \\
\text { Weight decay} & \multicolumn{2}{c}{10^{-4}} \\
\text { Max optimizer gradient norm} & \multicolumn{2}{c}{\text {0.5}} \\
\text { Target network update interval } & \multicolumn{2}{c}{400 \text { updates }} \\
\text { Value function rescaling } & \multicolumn{2}{c}{h(x)=\operatorname{sign}(x)(\sqrt{|x|+1}-1)+\epsilon x, \epsilon=10^{-3}}
\end{array}
$

\subsubsection{Zipf's Labyrinth}


$
\begin{array}{l|c}
\hline \text { Number of actors } & 256 \\
\text { Sequence length } & 128 (\text{prefix of } l=20 \text { burn-in}) \\
\text { Replay buffer size } & 10^{5} \\
\text { Minibatch size } & 32 \\
\text { Importance sampling exponent } & 0.6 \\
\text { Priority weight (max) } & 0.9 \\
\text { Exploration } \varepsilon & 10^{-3} \\
\text { Discount } \gamma & 0.997 \\
\text { Lambda-returns } \lambda & 0.8 \\
\text { Optimizer} & \text { AdamW }(\text { Loschilov \& Hutter, 2017) } \\
\text { Optimizer learning rate } & 3*10^{-4} \\
\text { Optimizer } \varepsilon & 1.25*10^{-6} \\
\text { Weight decay} & 10^{-4} \\
\text { Max optimizer gradient norm} & \text {0.5} \\
\text { Target network update interval } & 400 \text { updates } \\
\text { Value function rescaling } & h(x)=\operatorname{sign}(x)(\sqrt{|x|+1}-1)+\epsilon x, \epsilon=10^{-3}
\end{array}
$

\subsubsection{Zipf's Gridworld}

$
\begin{array}{l|c}
\hline  \text { Number of actors } & 256 \\
\text { Sequence length } & 128(\text {prefix of } l=3 \text { burn-in}) \\
\text { Replay buffer size } & 10^{5} \\
\text { Minibatch size } & 32 \\
\text { Importance sampling exponent } & 0.6 \\
\text { Priority weight (max) } & 0.9 \\
\text { Exploration } \varepsilon & 0.1 \\
\text { Discount } \gamma & 0.9 \\
\text { Lambda-returns } \lambda & 0.3 \\
\text { Optimizer } & \text { AdamW }(\text {Loschilov \& Hutter, 2017) } \\
\text { Optimizer learning rate } & 3*10^{-4}\\
\text { Optimizer } \varepsilon & 1.25*10^{-6} \\
\text { Weight decay} & 10^{-4} \\
\text { Max optimizer gradient norm} & \text {0.5} \\
\text { Target network update interval } & 10 \text { updates } \\
\text { Value function rescaling } & h(x)=\operatorname{sign}(x)(\sqrt{|x|+1}-1)+\epsilon x, \epsilon=10^{-3}
\end{array}
$

\section{Supplemental analyses} \label{apps:analyses}
\subsection{Train performance} \label{apps:analyses:train_perf}
Train performance, i.e. assessed on the Zipfian distributions that the agents encountered during training, was generally very high and obscured the relatively poor performance on rare scenarios. Train performance for Zipf's Labyrinth is reported directly in Table \ref{tab:labyrinth_results} under `All tasks (Zipfian)'.
\begin{table}[H]
\begin{tabular}{ll|l|ll|}
                                  &                       & \multicolumn{1}{c|}{Lifting task}                         & \multicolumn{2}{c|}{Putting task}                     \\
                                  &                       & \multicolumn{1}{l|}{All items (Zipf \(\alpha=3\))}    & \multicolumn{1}{l|}{Lifting (Zipf \(\alpha=2\))} & Putting (frequent only) \\ \hline
\multicolumn{1}{l|}{RL}           & IMPALA                & 0.999                      & 0.993                 & 0.987           \\
\multicolumn{1}{l|}{algorithm}    & V-MPO                  & 0.970                      & 0.791                 & 0.091      \\ \cline{1-2}
\multicolumn{1}{l|}{SSL}          & Vis recon             & 0.999                      & 0.992                 & 0.985           \\
\multicolumn{1}{l|}{(IMPALA +..)} & Vis+lang recon        & 0.999                      & 0.990                 & 0.976           \\ 
\multicolumn{1}{l|}{} & BYOL        & 0.999               & 0.989                 & 0.965           \\ \cline{1-2}
\multicolumn{1}{l|}{Memory}       & LSTM                  & 0.999                      & 0.993                 & 0.987           \\
\multicolumn{1}{l|}{system}       & Transformer           & 0.999                      & 0.991                 &  0.963         \\
\multicolumn{1}{l|}{(IMPALA +..)} & Transformer + recon           & 0.999              & 0.973                 &  0.801          \\
                                  & Prioritized (R2D2)    & \multicolumn{1}{l|}{1.000}                  & 0.995                 &      0.977
\end{tabular}
\caption{Training performance in Zipf's Playroom.}
\label{supp_tab:train_results:playroom}
\end{table}

\begin{table}[H]
\begin{tabular}{l|l|}

                        & All maps and objects (Zipf \(\alpha = 2\))        \\ \hline
IMPALA                  & 0.851                                             \\ \cline{1-2}
IMPALA + Vis recon      & 0.989                                             \\
IMPALA + BYOL           & 0.751                                             \\ \cline{1-2}      
R2D2                    & 0.672                                                \\
\end{tabular}
\caption{Training performance in Zipf's Gridworld. Median performance (across time) of median run for each condition.}
\label{supp_tab:gridworld_results}
\end{table}

\section{Learning curves} 
\label{apps:learning_curves}

\begin{figure*}
    \centering
    \begin{subfigure}[]{0.45\textwidth}
        \centering
        \includegraphics[width=\textwidth]{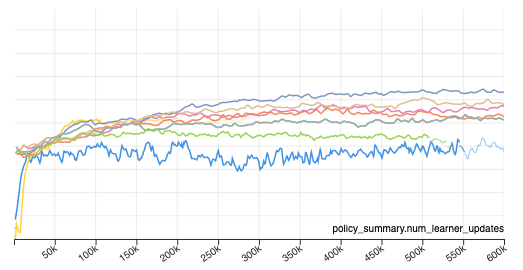}
        \caption{Rare - uniform.}
    \end{subfigure}
    \hfill
    \begin{subfigure}[]{0.45\textwidth}
        \centering
        \includegraphics[width=\textwidth]{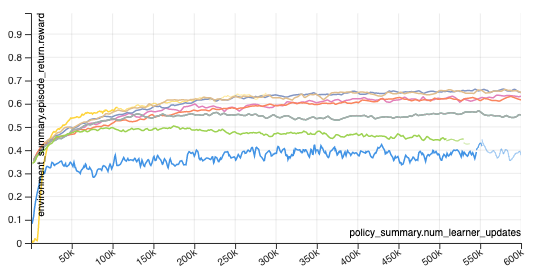}
        \caption{All - uniform.}
    \end{subfigure}
    \hfill
    \begin{subfigure}[]{0.45\textwidth}
        \centering
        \includegraphics[width=\textwidth]{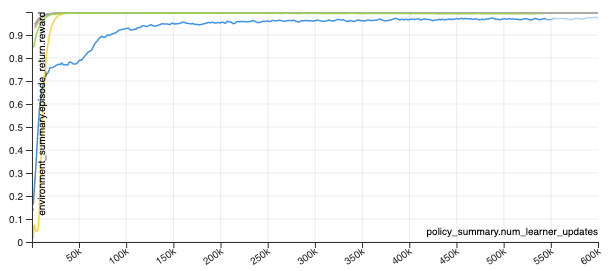}
        \caption{All - Zipfian (training distribution).}
    \end{subfigure}
    \hfill
    \begin{subfigure}[]{0.25\textwidth}
        \centering
        \includegraphics[width=\textwidth]{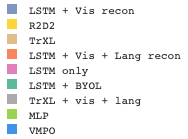}
    \end{subfigure}
    \caption{Learning curves for lifting task in Zipf's Playroom.}
    \label{apps:fig:learning_curves:lifting}
\end{figure*}

\begin{figure*}
    \centering
    \begin{subfigure}[]{0.45\textwidth}
        \centering
        \includegraphics[width=\textwidth]{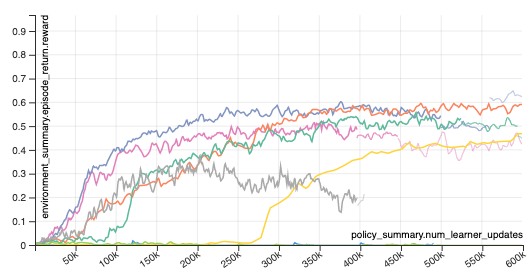}
        \caption{Rare - uniform.}
    \end{subfigure}
    \hfill
    \begin{subfigure}[]{0.45\textwidth}
        \centering
        \includegraphics[width=\textwidth]{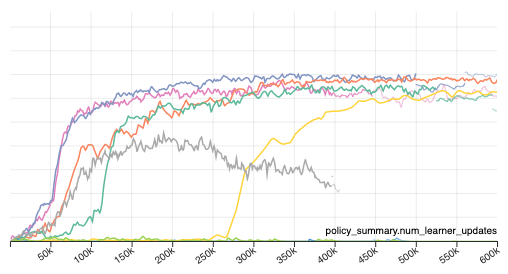}
        \caption{All - uniform.}
    \end{subfigure}
    \hfill
    \begin{subfigure}[]{0.45\textwidth}
        \centering
        \includegraphics[width=\textwidth]{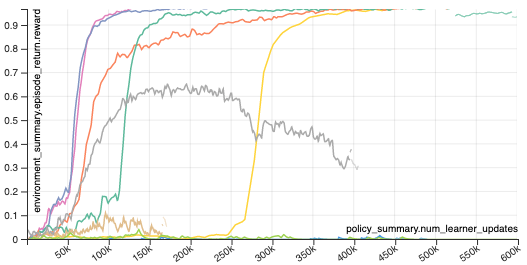}
        \caption{All - Zipfian (training distribution).}
    \end{subfigure}
    \hfill
    \begin{subfigure}[]{0.25\textwidth}
        \centering
        \includegraphics[width=\textwidth]{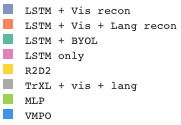}
    \end{subfigure}
    \caption{Learning curves for putting task in Zipf's Playroom.}
    \label{apps:fig:learning_curves:putting}
\end{figure*}

\begin{figure}
    \includegraphics[width=0.25\textwidth]{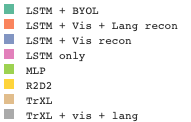}
    \caption{Learning curves for Zipf's Labyrinth (training on forward Zipf distribution).}
    \label{apps:fig:learning_curves:dmlab}
\end{figure}

\begin{figure*}
    \centering
    \begin{subfigure}[]{0.45\textwidth}
        \centering
        \includegraphics[width=\textwidth]{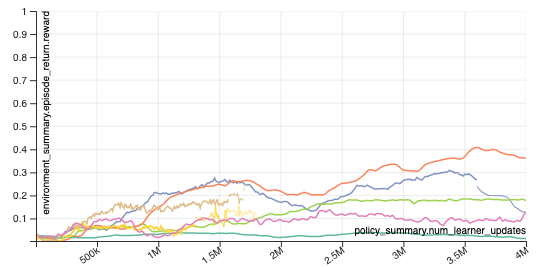}
        \caption{Rare - uniform.}
    \end{subfigure}
    \hfill
    \begin{subfigure}[]{0.45\textwidth}
        \centering
        \includegraphics[width=\textwidth]{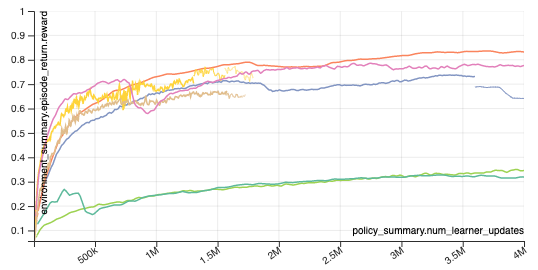}
        \caption{All - uniform.}
    \end{subfigure}
    \hfill
    \begin{subfigure}[]{0.45\textwidth}
        \centering
        \includegraphics[width=\textwidth]{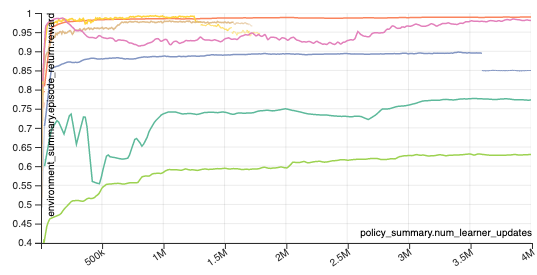}
        \caption{All - Zipfian (training distribution).}
    \end{subfigure}
    \hfill
    \begin{subfigure}[]{0.25\textwidth}
        \centering
        \includegraphics[width=\textwidth]{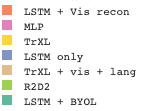}
    \end{subfigure}
    \caption{Learning curves for Zipf's Gridworld.}
    \label{apps:fig:learning_curves:gridworld}
\end{figure*}

Learning curves are displayed for Zipf's Playroom (Figs \ref{apps:fig:learning_curves:lifting}-\ref{apps:fig:learning_curves:putting}),  Zipf's Labyrinth (Fig \ref{apps:fig:learning_curves:dmlab}), and Zipf's Gridworld (Fig \ref{apps:fig:learning_curves:gridworld}).

\begin{figure}
    \centering
    \begin{subfigure}[]{\textwidth}
        \centering
        \includegraphics[width=\textwidth]{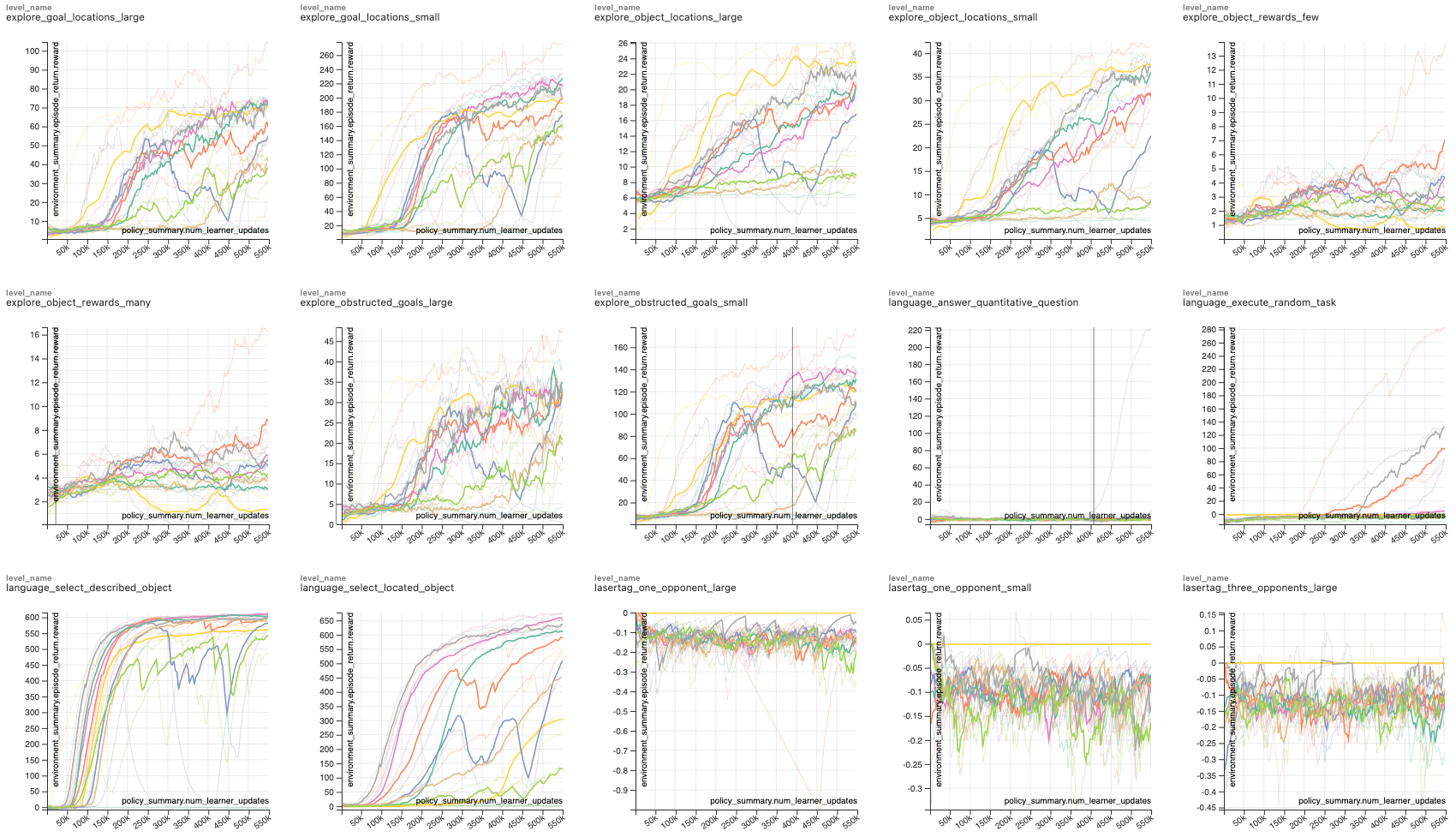}
    \end{subfigure}
    \hfill
    \begin{subfigure}[]{\textwidth}
        \centering
        \includegraphics[width=\textwidth]{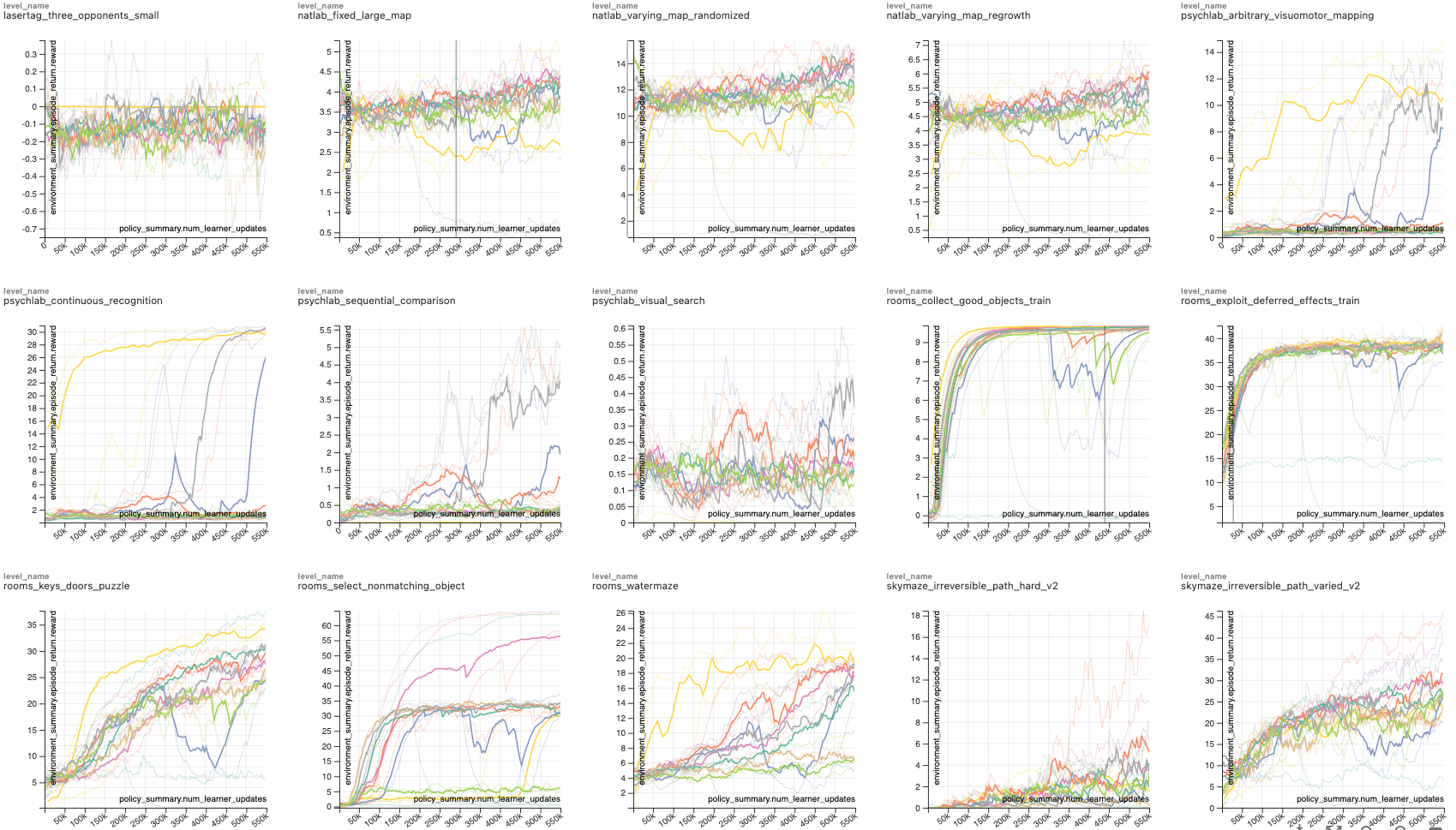}
    \end{subfigure}
\end{figure}

\end{document}